\documentclass{article}
\usepackage{spconf,amsmath,graphicx}
\usepackage{caption}
\usepackage{cite}
\usepackage{amsmath,amssymb,amsfonts}
\usepackage{algorithmic}
\usepackage{graphicx}
\usepackage{textcomp}
\usepackage{comment}
\usepackage{color}
\usepackage{multirow}

\title{LOCALIZING MOMENTS OF ACTIONS IN UNTRIMMED VIDEOS OF INFANTS WITH AUTISM SPECTRUM DISORDER}
\name{
Halil Ismail Helvaci\textsuperscript{\,1} \enspace
Sen-ching Samson Cheung\textsuperscript{\,1,2} \enspace
Chen-Nee Chuah\textsuperscript{\,2} \enspace
\textit{Sally Ozonoff\textsuperscript{\,3}} \enspace
\thanks{Research reported in this publication was supported by the National Institute of Mental Health of the National Institutes of Health under award number R01MH121344-01. The content is solely the responsibility of the authors and does not necessarily represent the official views of the National Institutes of Health.}}
\address{
\textsuperscript{\,1}Department of Electrical and Computer Engineering, University of Kentucky, Lexington, KY, USA\\
\textsuperscript{\,2}Department of Electrical and Computer Engineering, University of California, Davis, CA, US \\ 
\textsuperscript{\,3}UC Davis MIND Institute, University of California, Davis, CA, US \enspace \\
{\tt\small  halil.helvaci@uky.edu \enspace
\tt\small  sccheung@ieee.org \enspace
\tt\small  chuah@ucdavis.edu \enspace
\tt\small  sozonoff@ucdavis.edu \enspace}
}

\begin{document}
\maketitle
\begin{abstract}
Autism Spectrum Disorder (ASD) presents significant challenges in early diagnosis and intervention, impacting children and their families. With prevalence rates rising, there is a critical need for accessible and efficient screening tools. Leveraging machine learning (ML) techniques, in particular Temporal Action Localization (TAL), holds promise for automating ASD screening. This paper introduces a self-attention based TAL model designed to identify ASD-related behaviors in infant videos. Unlike existing methods, our approach simplifies complex modeling and emphasizes efficiency, which is essential for practical deployment in real-world scenarios. 
Importantly, this work underscores the importance of developing computer vision methods capable of operating in naturilistic environments with little equipment control, addressing key challenges in ASD screening. This study is the first to conduct end-to-end temporal action localization in untrimmed videos of infants with ASD, offering promising avenues for early intervention and support. We report baseline results of behavior detection using our TAL model. We achieve 70\% accuracy for look face, 79\% accuracy for look object, 72\% for smile and 65\% for vocalization.
\end{abstract}

\section{INTRODUCTION}
\label{sec:intro}

Autism Spectrum Disorder (ASD) is a prevalent developmental disorder that affects communication, social interaction, and behavior of children. The Centers for Disease Control and Prevention (CDC) reports that one in 36 children in the United States is diagnosed with ASD, with the prevalence of the disorder continuously increasing \cite{cdc_autism}. Children diagnosed with ASD demonstrate deficits in social communication and manifest distinctive patterns of attention towards both people and objects. Specifically, children with ASD often refrain from making eye contact in social interactions, and their eye gazes display distinct patterns that prominently differ from those observed in typical children \cite{ ronconi2018brief}. Families of children with ASD often experience stress, financial strain and dedicate substantial time to their care. Therefore, early diagnosis and interventions of ASD are crucial for the well-being of both the children and their families.

Research has shown that early interventions play a vital role in enhancing outcomes for children diagnosed with ASD \cite{zwaigenbaum2015early}. However, identifying ASD at an early stage poses a significant challenge, primarily attributed to limited access to professionals and the often overlooked symptoms of ASD. These limitations often lead to delays in treatment initiation, thereby missing the chance to improve treatment outcomes

The majority of ASD screening methods rely on ratings or questionnaires \cite{ozonoff2010prospective}, typically administered by trained professionals. However, this restricts their use in communities with limited access to healthcare professionals and public health resources. Additionally, these assessments are time-consuming, taking several hours, and children may need to undergo them at different ages for accurate diagnosis. Therefore, the development of efficient and automated ASD screening tools is imperative to alleviate the strain on healthcare infrastructure and reduce the burden on children. One promising avenue in this regard is to leverage machine learning techniques for the automated screening of ASD.Using ML, improving screening would be beneficial to getting children seen more quickly for diagnosis and intervention.

ML-based techniques have been utilized to formulate the diagnostic task as a classification problem. Previous research has explored this direction by predicting ASD using a ML facial recognition model that uses hand-crafted features \cite{nasser2019artificial}, face data \cite{wang2022asdface}, and brain images \cite{heinsfeld2018identification}. More recent research has shifted towards incorporating head movement \cite{sapiro2019computer}, raw images and facial landmarks\cite{healthcom}. However, these approaches typically analyze data for only a few seconds, at most, overlooking the temporal dynamics among behaviors.

The rising interest in video understanding has underlined the importance of effectively localizing moments of actions in untrimmed videos, or in other words, temporal action localization (TAL). In this text, untrimmed videos refer to raw unedited video footage that contains continuous sequence of frames without any cuts or edits. These videos often capture real-world scenarios or events without any modifications or alterations to their temporal structure. Different from action detection, TAL aims to identify the beginning and ending times of action instances along with the action labels. While TAL has proven successful on major benchmarks like THUMOS14 \cite{thumos14} and ActivityNet1.3 \cite{caba2015activitynet}, the models trained on public datasets lack appropriate actions for analyzing ASD behaviors. The behaviors essential for diagnosing ASD, described in detail in Section \ref{dataset collection}, diverge from those observed in public datasets, which typically include activities like diving, biking, and high jumps.

Existing action detection methods typically entail complex modeling, custom loss functions, anchor design, and external post processing \cite{zhang2022actionformer, xu2020g, liu2022end}. In this study, we demonstrate that TAL can accurately and efficiently detect ASD related behaviors in untrimmed videos. We develop a simple self-attention based model for TAL, motivated by the idea that a localization strategy could capture temporal behavior relationships reducing the burden of hand labeling by triaging relevant portions of the video.

Deep learning models face a challenge due to their extensive parameters, necessitating large labeled datasets. This issue is particularly pronounced in medical domains, where transfer learning can struggle with diverse problems \cite{xiao2018opportunities}. While early attempts at ASD diagnosis with deep learning show promise, they rely on small datasets, raising concerns about the generalization
of the developed techniques. In this study, our interdisciplinary team presents a framework that employs a computer vision and deep learning to identify autism related behaviors from videos. The technical advancement utilizes a meticulously curated large video dataset that focuses on children's social interactions, gathered at the Medical Investigation of Neurodevelopmental Disorders (MIND) Institute at the University of California, Davis. Our machine learning approach involves a pipeline employing TAL to predict actions from raw, untrimmed videos. In adapting TAL to the specific application of localizing behaviors in videos of children with ASD, we encounter unique challenges and opportunities. Our approach aims to address the complexity of detecting behaviors such as 'look face' and 'smile,' which often exhibit rapid and subtle temporal characteristics. Moreover, challenges inherent in adapting TAL models to videos of children with ASD include variability in individual behavior patterns, and the potential presence of confounding factors such as non-verbal cues or repetitive movements.

In this paper, we introduce the following contributions to the field of human behavior detection with a focus on children with ASD. 

\begin{enumerate}
    \item We introduce an end-to-end self-attention based temporal action localization model to automatically identify ASD relevant behaviors exhibited by infants.
    \item To the best of our knowledge, this study is the first to conduct end-to-end temporal action localization of behaviors in untrimmed videos featuring infants with ASD.
\end{enumerate}

\section{RELATED WORK}
\label{sec:related work}

Significant progress has been made in utilizing computer vision and machine learning to analyze children's behaviors in autism research. One example is ResearchKit, an iPhone app developed by the Sapiro Lab at Duke University that utilizes social and non-social videos to capture head pose and emotions for each subject \cite{sapiro2019computer}. Another study used a facial expression recognition software to determine the level of engagement of children in response to video stimuli, by categorizing their expressions as neutral, positive or negative \cite{hashemi2018computer}. Authors in \cite{li2019facial} introduced an end-to-end system that leverages face images to train a neural network . More recent efforts have centered on extracting raw images and facial landmarks from videos to make frame-level action predictions, showing success in identifying behaviors related to ASD \cite{healthcom}. 

Convolutional neural networks (CNNs) have been a common choice due to their ability to capture spatial and temporal information within a video sequence. A recent study introduced an action recognition model that employs a 3 dimensional (3D) CNN network to extract features from both the spatial and temporal domains, using 3D convolutions \cite{li2020classifying}. On the other hand, another research presented a CNN-LSTM network, where the CNN is responsible for capturing the spatial features and the LSTM is used for capturing the temporal features \cite{ullah2017action}. However, a limitation of both methods is their reliance on trimmed videos, with model architectures designed to processing only short durations of videos, typically a few seconds at most.

As a fundamental task in video understanding, TAL is a challenging task that involves identification of actions in untrimmed videos. In contrast to trimmed videos, untrimmed videos can encompass seconds or even minutes of content before or after the action occurs. Current state of the art TAL methods can be mainly categorized into three groups: anchor-free, anchor-based and query-based. Notably,  ActionFormer \cite{zhang2022actionformer}, G-TAD \cite{xu2020g} and TadTR \cite{liu2022end} are the state of the art representatives for each method, respectively.

ActionFormer is an anchor-free method that combines multi-scale feature representation with local self-attention that classifies every moment in time while estimating the corresponding action boundaries \cite{zhang2022actionformer}. Specifically, ActionFormer extracts a feature pyramid from videos using a multi-scale transformer encoder that is connected to classification and regression heads allowing the model to produce action candidates at each time step. The key contribution lies in the implementation of a single stage anchor-free transformer network that outperforms previous methods \cite{xu2020g, liu2022end} on THUMOS14 by a large margin and surpassing previous methods when using identical features on ActivityNet1.3. 

G-TAD is an anchor-based method that utilizes videos as graphs and all snippets in each video as a graph node \cite{xu2020g}. The main idea is to capture the context of the video by applying graph convolutions on the nodes. Each node is sampled as a potential action boundary and paired nodes become anchors. SG-Align is used for extracting the paired features within the temporal region of each anchor. Lastly, the anchors are classified by fully connected layers upon the aligned features. Even though G-TAD has state of the art results for anchor-based models, the performance is subpar compared to anchor-free and query-based methods. The main constraint of this method is the pre-determined sizes of the anchors, while the events can last from seconds to even minutes. 

TadTR is a query-based method that uses a transformer-based model which maps learned action query embeddings to matching action predictions using the full transformer encoder-decoder architecture \cite{liu2022end}. The features are supplied to the transformer encoder, which models the long-range dependency of the temporal dimension. The decoder corresponds each query to the global context using cross-attention that predicts action classes and boundaries. Unlike the ActionFormer, TadTR does not utilize non-max supression (NMS) \cite{bodla2017soft} as a post processing step. To improve the confidence scores and action boundaries, TadTR uses segment refinement that iteratively refines the boundaries in each decoder layer and re-calculates the confidence score according to the final predicted boundaries. However, even though the model utilizes a transformer architecture, the refinement process is not purely based on transformers, which performs inferior to a complete transformer-based model, ActionFormer.

\section{DATASET COLLECTION}
\label{dataset collection}

This study is built upon a collection of videos from the Infant Sibling Study, conducted at the UC Davis MIND Institute. The study, approved by the Institutional Review Board (IRB), aims to track the development of autism symptoms in infants over their first three years. Evaluations were conducted every six months, starting at 6 months and ending at 36 months. Infants were categorized into two groups: High Risk and Low Risk. The High-Risk group comprises infants with an older sibling diagnosed with ASD based on both the Autism Diagnostic Observation Schedule (ADOS) and the Social Communication Questionnaire (SCQ). Further information regarding the study's inclusion and exclusion criteria can be found in \cite{ozonoff2010prospective}.

The video dataset comprises 1707 recordings featuring 365 infants participating in various adult-child play tasks. Each recording captures a 3-minute interactive play session involving either an examiner or a parent. Given that ASD symptoms often manifest through social behaviors such as eye contact, play activities, and communication with others, our experiments are tailored to observe how children respond during interactions with examiners. These interactive sessions take place in a camera-equipped room where the actions of both examiners and children are recorded. Researchers, positioned behind a private glass window, have a comprehensive view of all activities and oversee the recording process. During these experiments, a trained examiner, knowledgeable in assessing autistic behaviors, engages face-to-face with the child while the parent may be present nearby. The examiner uses various toys like dolls or small cellphones to engage the child and may introduce interruptions with different stimuli to assess the child's responsiveness. All interactions and responses from the child are recorded. Each recording setup includes a camera to monitor the child's actions and another to capture the examiner's interactions. For each visit, three play sessions can be documented at most.

\begin{figure}[!h]
\centering
\begin{tabular}{@{\hspace{0mm}}c@{\hspace{0.5mm}}c@{\hspace{0.5mm}}c}
\fbox{\includegraphics[width=0.45\columnwidth, height=0.4\columnwidth]{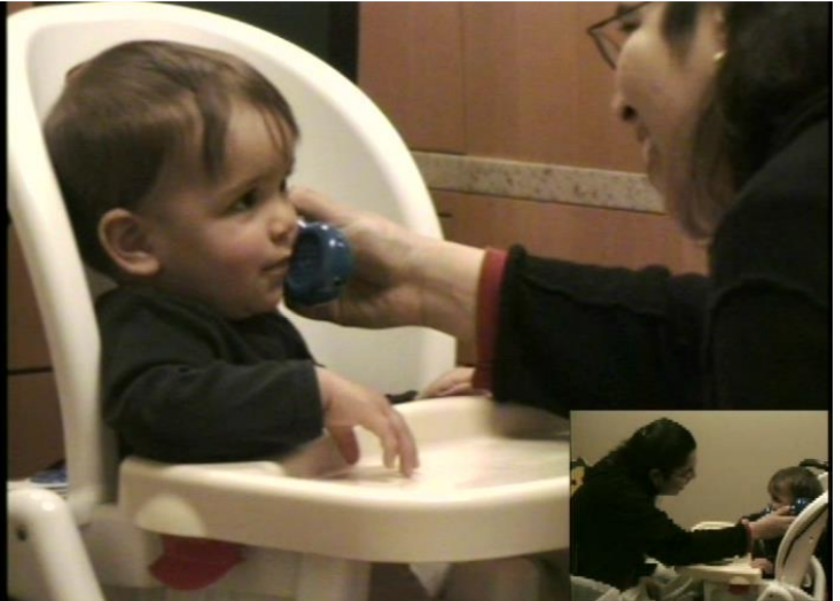}}
& \fbox{\includegraphics[width=0.45\columnwidth, height=0.4\columnwidth]{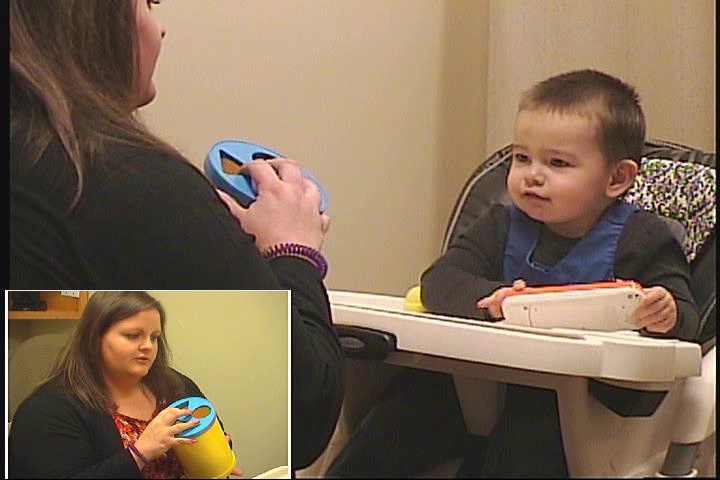}}\\
{(a) Look Face} & {(b) Look Object}\\

\fbox{\includegraphics[width=0.45\columnwidth, height=0.4\columnwidth]{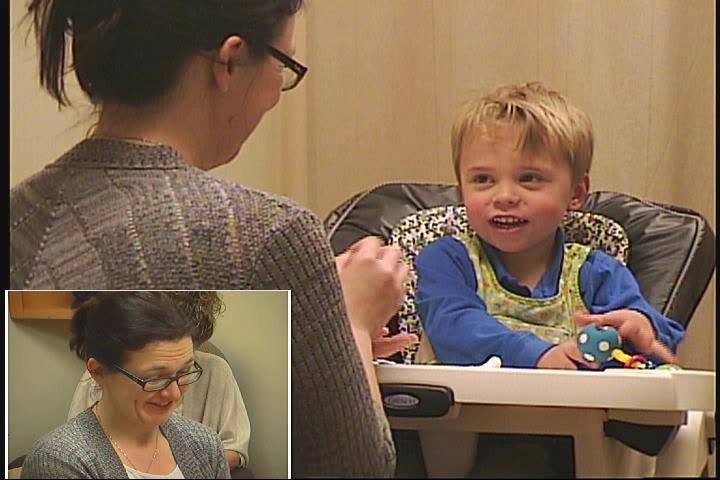}}
& \fbox{\includegraphics[width=0.45\columnwidth, height=0.4\columnwidth]{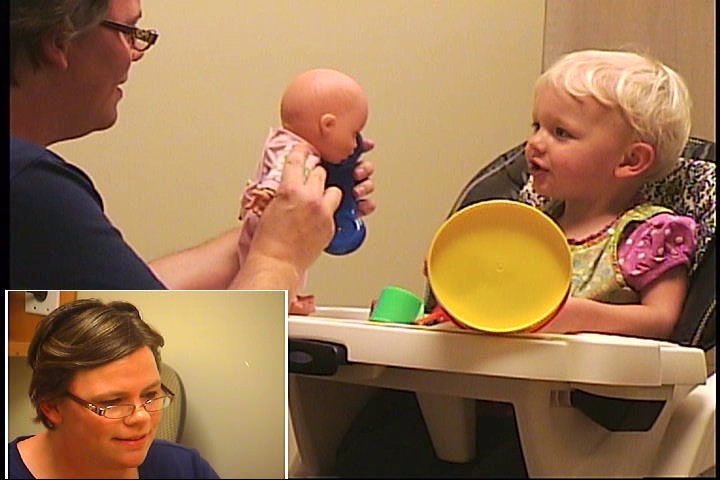}}\\
{ (c) Smile} & { (d) Vocal}\\
\end{tabular}
\caption{Visualization of the four behaviors of interest}
\label{fig:autism_data_example}
\end{figure}

For each video, our analysis focuses on four distinct actions: 1) "Look face" indicating whether the subject makes eye contact with the partner by looking at their face; 2) "Look object" assessing whether the subject looks at an object of interest; 3) "Smile" determining whether the subject is exhibiting a smile; and 4) "Vocal" examining whether the subject produces sounds or speaks. Instances of each action is illustrated in Fig~\ref{fig:autism_data_example}.

The labels that are used for our supervised learning were manually coded and labeled by trained clinical coders. All of the video recordings were carried out using a behaviotal observation softwared called Observer 5.0, developed by Noldus. Coders received an initial training to have agreement of at least 90\% on all codes. Moreover, to ensure reliability, double coding by master coders was performed on 15\% of the data, resulting in highly favorable intra-class correlation coefficients (ICC) for all codes (gaze to face: 0.95, gaze to object: 0.98, smile: 0.96, and vocalization: 0.93).

In total, we have 1707 videos that are annotated. For the initial investigation that is discussed in this study, we utilize 349 videos of 59 subjects. We split the videos into two: train set and a test set. To ensure the generalization of the trained models, there is no overlap of human subjects in the train and test set. The detailed statistics can be found in Table \ref{table:data_stat}

\begin{table}[]
\captionsetup{justification=centering, font=small}
\caption{ STATISTICS OF TRAIN AND TEST DATASETS}
\centering
    \begin{tabular}{c|c|c|c}
    \hline
     & \#Children & \#Videos & Video Percentage \\ \hline
     Train & 59  & 183 & 52.4\%  \\ \hline
     Test & 60  & 166 & 47.6\%  \\ \hline
    \end{tabular}
    \label{table:data_stat}
\end{table}

\section{LOCALIZING MOMENTS OF ACTIONS IN UNTRIMMED VIDEOS USING DEEP LEARNING}
\label{sec:method}



In this work, the untrimmed input video \textbf{X} is denoted by a collection of feature vectors \textbf{X} $ = \{ x_1, x_2, ..., x_t\}$, where $\{ 1, 2, ... , t\}$ represents the discrete time intervals, and $t$ refers to the entire duration of the video. The TAL task, as described in literature \cite{zhang2022actionformer,xu2020g, liu2022end}, involves generating behavior labels \textbf{Y} $ = \{y_1, y_2, ..., y_N\}$ based on an input video $X$, where $N$ represents the number of behavior instances observed in a video. It is important to note that $N$ is not a fixed value and may differ for each video. Each instance consists of it's start time $S_i$, end time $E_i$ and behavior label $C_i$, where an instance is represented as $Y_i = (S_i, E_i, C_i)$ and $S_i < E_i$, $S_i \in [1,i]$, $E_t \in [1,i]$, $C_i \in \{1, 2, ..., C\}$ with $i\in{1,2,\ldots,N}$, which denotes the set of the pre-defined action classes. Contrary to previous methods \cite{xu2020g, liu2022end}, our method aims to predict the probability of an action $p(C_t)$ being present at any given moment $t$ and further regress on the time differences $D^s_t$ and $D^e_t$ between $t$ and the respective start time and end time. This reformulates $y_i = (s_i, e_i, C_i)$ into 
\begin{equation}
    y_t = (D^S_t, D^E_t, p(C_t))
\end{equation}

For each moment t captured in the video X, the formula evaluates it as a potential action and calculates the probability of it being an action. Furthermore, the formula determines the distance between the current moment and the start and end points of the action boundaries. The predictions are later decoded from $y_t = (D^s_t, D^e_t)$ to yield $S_t = t - D^s_t$, $E_t = t +D^e_t$ where $S_t$ and $E_t$ are action start and end times in seconds respectively.

\begin{figure}[!t]
    \centering
    \includegraphics[width= 1.0\linewidth]{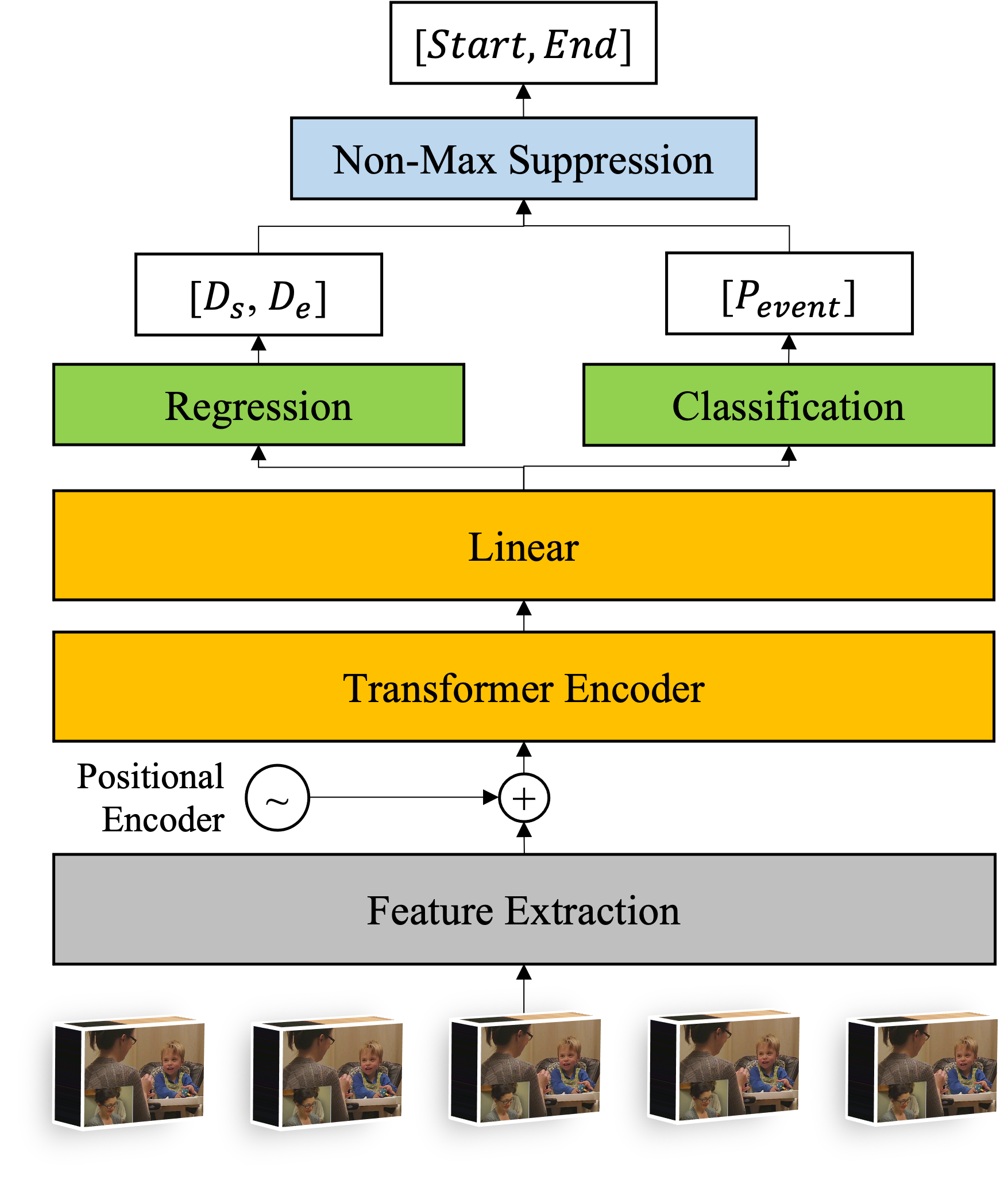}
    \captionsetup{justification=centering}
    \caption{Visualization of the End-to-End Behavior Localization Pipeline}
    \label{fig:halil_pipeline}
\end{figure}

Our method is similar to other deep learning techniques such that it follows the format of $f(X) \rightarrow Y$, but instead of relying on traditional CNN and Recurrent Neural Network (RNN) based methods, it employs a transformer-based network \cite{vaswani2017attention} as its $f$ function. Specifically, our model is composed of a positional encoder, transformer encoder, linear layers followed by a classification and regression head. After generating action candidates, Soft-NMS \cite{bodla2017soft} is applied to eliminate instances that have a high degree of overlap, resulting in the production of final output behaviors. The complete model is demonstrated in Figure \ref{fig:halil_pipeline}. A comprehensive breakdown of each step is outlined below.

\subsection{Feature Extraction}

Raw videos are typically high-dimensional and contain irrelevant information that may complicate the processes of action recognition and localization. Feature extraction serves as a means to reduce data complexity by isolating pertinent information related to behavior detection. Notably, pre-training models on the Kinetics dataset has shown significant enhancement in performance \cite{zhang2022actionformer,xu2020g, wang2022asdface}. Inspired by this observation, we employ a two-stream I3D model pre-trained on the Kinetics dataset for feature extraction
from the ASD videos. Specifically, we feed $64$ consecutive frames into the I3D model without any overlap, resulting in the extraction of 1024 features for both the RGB and Flow streams. Subsequently, we concatenate these features to create a tensor vector with $2048$ features. The extracted features then undergo a positional encoding procedure, following the encoding scheme as $PE_{(pos, 2i)} = \sin(pos/10000^{2i/d_{\text{model}}})$ and $PE_{(pos, 2i+1)} = \cos(pos/10000^{2i/d_{\text{model}}})$ \cite{vaswani2017attention}, where $pos$ is the segment position and $ i \in \{0,...,2037\}$ 

\subsection{Feature Encoding Using Transformer Encoder}

After positional encoding is applied to the input features, they are passed into the transformer encoder network, which consists of a combination of the self-attention mechanism \cite{vaswani2017attention} followed by a feedforward layer. Self-attention computes the relationship between each element in the input sequence, resulting in a weighted representation of the sequence. This process captures complex dependencies within the input sequence by comparing each positional input with others. In mathematical terms, self attention extracts key ($K$), query ($Q$) and values ($V$) for a given input ($I$), with a fixed dimension ($A$) and time step ($t$) are calculated as
\begin{equation}
    Q = I W_Q, \enspace K = I W_K, \enspace V = I W_V \enspace
\end{equation}

\noindent where $W_Q \in \mathbb{R}^{A \times A_Q}$, $W_K \in \mathbb{R}^{A \times A_K}$ and $W_V \in \mathbb{R}^{A \times A_V}$. The output of self-attention is then determined as

\begin{equation}
    \sigma(Q \times K^t / \sqrt{A_q}) \times V
\end{equation} 

\noindent with $\sigma$ representing the softmax function. The self attention layer is followed by a linear layer to output the encoded features. In our case, we utilize a multi-head self-attention (MSA) mechanism \cite{vaswani2017attention}, which involves running multiple self-attention operations in parallel.

More precisely, our transformer encoder includes 16 attention heads with a internal feed forward layer dimension of 168. Layer normalization is applied after each MSA and MLP followed by residual connections after every block. The MLP layer utilizes GELU activation. Our model uses the entire duration of the video for MSA calculations, and does not limit the attention span to individual segments or a short window.

\subsection{Decoding Actions}

Next, we implement our decoder network which consists of two separate heads, one for classification and the other for regression. Each linear layer has a hidden layer dimension of 1024. Following each layer, a 1D batch normalization is applied, with ReLU activation and an output shape of 512. The classification head comprises of a fully connected layer, with an output dimension of the number of action classes, in our case 2, since we are running our models separately for each behavior class.  A Softmax activation is attached to the output to predict the probability of an event $P_{event}$ occurring at each time step. Similar to the classification head, the regression head evaluates each time step and performs distance predictions $D^s, D^e$ of when an action begins and ends. The regression head is a fully connected layer, with an output size of 2, $D^s$ and $D^e$. Lastly, a Soft-NMS \cite{bodla2017soft} is implemented to remove highly overlapping instances, leading to the final outputs of behaviors. The NMS algorithm takes a combined vector $[D_s, D_e, P_{event}]$ as its input. It selects the proposals with the highest confidence scores and checks for overlaps with other bounding boxes. If the overlap exceeds a pre-defined threshold of 0.5, the algorithm discards the overlapping box.

During training, we employ two distinct loss functions. We use Mean Squared Error for the regression network. For the classification network, we utilize focal loss \cite{lin2017focal}, which is suitable to handle significant imbalance between behavior and non-behavior instances. To optimize the model, we combine the focal loss and regression loss to form the total loss. The model is trained for 100 epochs, with a batch size of 10 and a learning rate $10^{-3}$, which is  decayed by a factor of 0.01 when the combined loss does not improve for 5 epochs. Stochastic gradient descent (SGD) is used as the optimization function. 

During inference, the entire duration of the sequences are fed into the model. Our model takes the video input \textbf{X} and outputs $y_t = (D^S_t, D^E_t, p(C_t))$ which is later decoded into seconds $S_t = t - D^s_t$, $E_t = t +D^e_t$ where $S_t$ and $E_t$ are behavior start and end times. Because of the notable imbalance between instances of behavior and non-behavior, we adjust the decision threshold during inference to 0.4 instead of the default 0.5 used with Softmax.

\section{EXPERIMENTAL RESULTS}
\label{sec:experimental results}

To demonstrate the efficacy of our TAL method in identifying the four behaviors, we deploy a base classification network that employs the identical model structure as the TAL approach, with the exception of the regression head. Based on the counts of true positives (TP), true negatives (TN), false positives (FP), and false negatives (FN), the following metrics are commonly employed in literature: accuracy, defined as $=\frac{TP+TN}{TP+FP+TN+FN}$; sensitivity or recall, calculated as $= \frac{TP}{FN+FP}$; specificity, represented as $= \frac{TN}{FP+TN}$; precision, computed as $=\frac{TP}{TP+FP}$; and F1 score, given by $=\frac{2PR\cdot RE}{PR+RE}$. Accuracy finds prevalent use in machine learning literature, while sensitivity and specificity are prominent in medical literature. 

\begin{table}[!h]
\caption{RESULTS WITH CLASSIFICATION LOSS}
\centering
\small 
    \begin{tabular}{|c|c|c|c|c|} 
    \hline
              & Sensitivity & Specificity & F1-score & Accuracy \\ \hline
    Look Face & 0.34        & 0.79       & 0.57     & 67\%     \\ \hline
    Look Object & 0.85     & 0.50       & 0.68    &75 \%        \\ \hline
    Smile     & 0.36       & 0.83        & 0.55    & 67\%     \\ \hline
    Vocal    & 0.45        & 0.68        & 0.55     & 62\%     \\ \hline
    \end{tabular}
\label{table:classification-results}
\end{table}

Table~\ref{table:classification-results} showcases the preliminary outcomes obtained from training solely with the classification head. The base classifier demonstrates strong initial performance for look object with 75\% accuracy. However, the performance of the model is poor when identifying look face, smile and vocal with all having accuracy's below 67\% and F1 scores below 57\%. Moreover, we demonstrate that our TAL model enhances these results, as depicted in Table~\ref{table:tal-results}. The new accuracy's for the three actions surpass 70\%, while the F1-scores exceed 58\%. Nevertheless, for both models, the ability to detect vocalization lags behind due to the absence of audio data in the detection process. We are currently exploring the integration of both video and audio inputs into a unified framework. 

\begin{table}[!h]
\caption{RESULTS WITH COMBINED LOSS}
\centering
\small 
    \begin{tabular}{|c|c|c|c|c|} 
    \hline
              & Sensitivity & Specificity & F1-score & Accuracy \\ \hline
    Look Face & 0.39        & 0.81        & 0.60     & 70\%     \\ \hline
    Look Object & 0.90      & 0.45        & 0.70     & 79\%     \\ \hline
    Smile     & 0.37        & 0.86       & 0.58    & 72\%     \\ \hline
    Vocal     & 0.56        & 0.57        & 0.56     & 65\%     \\ \hline
    \end{tabular}
\label{table:tal-results}
\end{table}

Table~\ref{table:map-results} presents the average precision scores at various t-IoU (temporal Intersection over Union) thresholds for the TAL task. The table displays the performance metrics for the four behaviors mentioned above. Each row represents a specific action, while each column corresponds to a different t-IoU threshold ranging from 0.1 to 0.7, with an additional column showing the average across all thresholds. Notably, the look object action demonstrates consistently high average precision scores across different thresholds, indicating robust performance in temporal localization of this action. Conversely, the actions look face, smile and vocal shows a declining trend in average precision scores as the t-IoU threshold increases. Overall, the table provides valuable insights into the performance of the model for different actions and t-IoU thresholds, facilitating further analysis and optimization for our proposed TAL approach.

\begin{table}[!h]
\caption{AVERAGE PRECISION RESULTS FOR TAL}
\centering
\small 
\setlength{\tabcolsep}{10pt}
    \begin{tabular}{|c|c|c|c|c|c|} 
    \hline
              & 0.1 & 0.3 & 0.5 & 0.7 & Avg. \\ \hline
    Look Face   & 0.95       & 0.74        & 0.45     & 0.13     & 0.58    \\ \hline
    Look Object & 0.99       & 0.97        & 0.89     & 0.77     & 0.91    \\ \hline
    Smile       & 0.97       & 0.77        & 0.48     & 0.15     & 0.60      \\ \hline
    Vocal       & 0.85       & 0.67        & 0.52     & 0.40     & 0.61   \\ \hline
    \end{tabular}
\label{table:map-results}
\end{table}
 
\section{CONCLUSIONS}
\label{sec:conclusions}

In this paper, we presented a comprehensive video dataset capturing children's social interactions, annotated with behaviors crucial for ASD diagnosis. We proposed a machine learning framework tailored to this dataset, comprising of a behavior localizing model leveraging raw, unedited videos. For behavior detection, we introduced a baseline temporal action localization approach. Our method involved end-to-end training on raw video videos. Looking ahead, our future work aims to incorporate audio cues for vocalization detection, and implement self-supervised methods for identifying behaviors to refine look-face and look-object detection capabilities. Additionally, identifying all the behaviors simultaneously is a challenging task that we intend to investigate in the future. 

\bibliography{refs}

\end{document}